\documentclass[journal ]{new-aiaa}
\usepackage[utf8]{inputenc}
\usepackage{textcomp}

\usepackage{graphicx}
\usepackage{amsmath}
\usepackage[version=4]{mhchem}
\usepackage{siunitx}
\usepackage{longtable,tabularx}
\usepackage{subfigure}
\usepackage{diagbox}

\title{Fusion Detection via Distance-Decay IoU and weighted Dempster-Shafer Evidence Theory}

\author{Fang Qingyun \footnote{Graduate Research Assistant. } and Wang Zhaokui\footnote{Associate Professor, Department Head.  Member AIAA}}
\affil{Department of Aeronautics and Astronautics Engineering, Tsinghua University, Beijing, China, 100084}

\begin{document}

\maketitle

\begin{abstract}
In recent years, increasing attentions are paid on object detection in remote sensing imagery. However, traditional optical detection is highly susceptible to illumination and weather anomaly. It is a challenge to effectively utilize the cross-modality information from multi-source remote sensing images, especially from optical and synthetic aperture radar images, to achieve all-day and all-weather detection with high accuracy and speed. Towards this end, a fast multi-source fusion detection framework is proposed in current paper. A novel distance-decay intersection over union is employed to encode the shape properties of the targets with scale invariance. Therefore, the same target in multi-source images can be paired accurately. Furthermore,  the weighted Dempster-Shafer evidence theory is utilized to combine the optical and synthetic aperture radar detection, which overcomes the drawback in feature-level fusion that requires a large amount of paired data. In addition, the paired optical and synthetic aperture radar images for container ship Ever Given which ran aground in the Suez Canal  are taken to demonstrate our fusion algorithm. To test the effectiveness of the proposed method, on self-built data set, the average precision of the proposed fusion detection framework outperform the optical detection by 20.13\%.
\end{abstract}

\section*{Nomenclature}


{\renewcommand\arraystretch{1.0}
\noindent\begin{longtable*}{@{}l @{\quad=\quad} l@{}}

$\operatorname{Bel}$ & belief function \\
$ D $ & absolute compatibility coefficient \\
$ m $ &  mass function \\
$ \mathscr{P} $ & power set \\
$\operatorname{pl}$ & plausibility function \\
$ R $ &  relative compatibility coefficient \\
$ s $ & similarity \\
$ w $ &  weight of the hypothesis \\

$\Theta$ & frame of discrimination \\


\end{longtable*}}

\section{Introduction}

\lettrine{I}{n} recent years, multi-source high-resolution remote sensing data has been continuously enriched, and the implementation of fast and accurate object detection in all weather and all day on satellites will become one of the important directions of satellite remote sensing technology in the future. Object detection not only determines the class of interest but also gives the location of each prediction. This technology plays a key role in civilian and military fields such as port and airport flow monitoring, traffic diversion, and finding lost ships.

Due to the long-term development of deep learning, especially convolutional neural networks(CNNs), object detection has yielded a series of remarkable achievements, such as pedestrian and vehicle detection in autonomous driving. Currently, there are also many methods based on deep learning for object detection in remote sensing imagery, like the general object detection, they can be broadly classified into two-stage and one-stage. The two-stage detector divides the detection into two stages, localization and recognition, and has an additional step of region proposal compared to the single-stage detector. Essentially, the region proposal is equivalent to coarse-grained detection. And then fine detection is performed by the detection head, thus it generally has higher detection accuracy, such as R-CNN\cite{girshick2014rich} , Fast R-CNN\cite{girshick2015fast} , Faster R-CNN\cite{ren2015faster}  and other detectors\cite{cai2018cascade,li2019scale,pang2019libra}. In addition, single-stage detectors, such as RetinaNet \cite{lin2017focal}, SSD\cite{liu2016ssd}, EfficientDet \cite{tan2020efficientdet}, YOLO\cite{redmon2016you}, etc., utilize regression to achieve localization and recognition simultaneously. Compared with the two-stage method, they pursue a balance of accuracy and speed, and because of the single-stage, the network structure is relatively concise, and the requirements for the device are relatively reduced.

With the rapid development of remote sensing technology, it is possible to use satellite formations to achieve simultaneous multi-source imaging of the same region. UrtheCast Technology company plans to build, the world's first fully-integrated, multispectral optical and Synthetic Aperture Radar (SAR) commercial constellation of Earth Observation satellites known as the OptiSAR™ Constellation. By flying the satellites in tightly-paired SAR and optical tandem formations, the constellation will provide unmatched space-imaging capabilities\cite{tyc2017rich}.  
However, most of the proposed object detection methods are designed for single-source remote sensing data, which has limitations. Although optical images are rich in texture and color information, they are affected by illumination and weather anomaly, for example, they cannot work at night and in rainy, snowy, and foggy weather. SAR satellites can work in all day and all weather, but the image carries a small amount of information. If the complementary information between multi-source images is fully fused, a more objective and comprehensive interception of the target can be obtained to facilitate visual interpretation and computer processing.

Compared with object detection in single-source images, object detection in multi-source images has just started, and most of the papers are aimed at 3D object detection and pedestrian detection in autonomous driving\cite{liang2019multi,liang2018deep,wolpert2020anchor,zheng2019gfd,zhou2020improving,zhang2019weakly}. However, there are few related papers on detection in multi-sensor remote sensing imagery\cite{wu2020vehicle}\cite{schilling2018object}, and the proposed object detection networks are large in scale and slow in speed, which is not suitable for deployment on satellites in the future. In summary, we think the main challenges to achieving online multi-source detection on satellites are as follows:

\begin{enumerate}
\item  Lack of paired multi-sensor remote sensing image data sets, it is difficult to obtain multi-source paired images, and the process of manual production of paired data is complicated and costly.
\item The shapes of targets in multi-sensor images are quite different. For example, the shapes of ships in optical and SAR images are completely different. This leads to the difficulty of pairing the same targets between different source images.
\item  The power and computing power of satellite platforms are limited, and the limited computing power will limit the accuracy, timeliness and complexity of detection algorithms.
\end{enumerate}

In order to overcome the above three challenges,  we investigate the impact and explore solutions in the current paper. We construct a ship data set with 76 pairs of optical and SAR images, including 1152 ship targets. Note that the 76 pairs of images are only utilized as a test set. Further, a new fast detection  method base on Distance-Decay Intersection over Union (DDIoU) and decision-level fusion in multi-sensor remote sensing imagery is proposed.  The research of multi-sensor detection mostly adopts the idea of feature-level fusion \cite{zhou2020improving,liang2019multi}, while it requires a large amount of paired data for the network to learn cross-modality features during the training process. 
However, using decision-level fusion can greatly reduce the reliance on paired data because it eliminates the need during the  training process. It only uses paired multi-source images when testing, which makes this approach have natural advantages on more occasions than feature-level fusion. The DDIoU is designed to solve the problem of pairing the same targets between different source images due to its efficient coding ability. Furthermore, to handle the limited computing power problem, we adopt a new lightweight fusion algorithm called weighted D-S evidence theory, which takes only 4-5 milliseconds to run. Additionally, the proposed method could have a wilder application not only for remote sensing imagery, i.e., for multispectral object detection in general images\cite{chen2021multimodal}.

The rest of this paper is organized as follows. Our entire fusion detection framework is given in section II. The experimental results and discussions are presented in section III. Finally, the conclusions are drawn in section IV.

\section{Fusion Detection Framework}

\subsection{Algorithm Structure}

In order to combine the complementary information between the multi-sensor images, this paper adopts a decision-level fusion detection method to obtain a more objective and comprehensive interpretation of the targets, so as to improve the performance.

\begin{figure}[!htbp]
	\centering
	\vspace{0.2in}
	\includegraphics[width=0.35\textwidth]{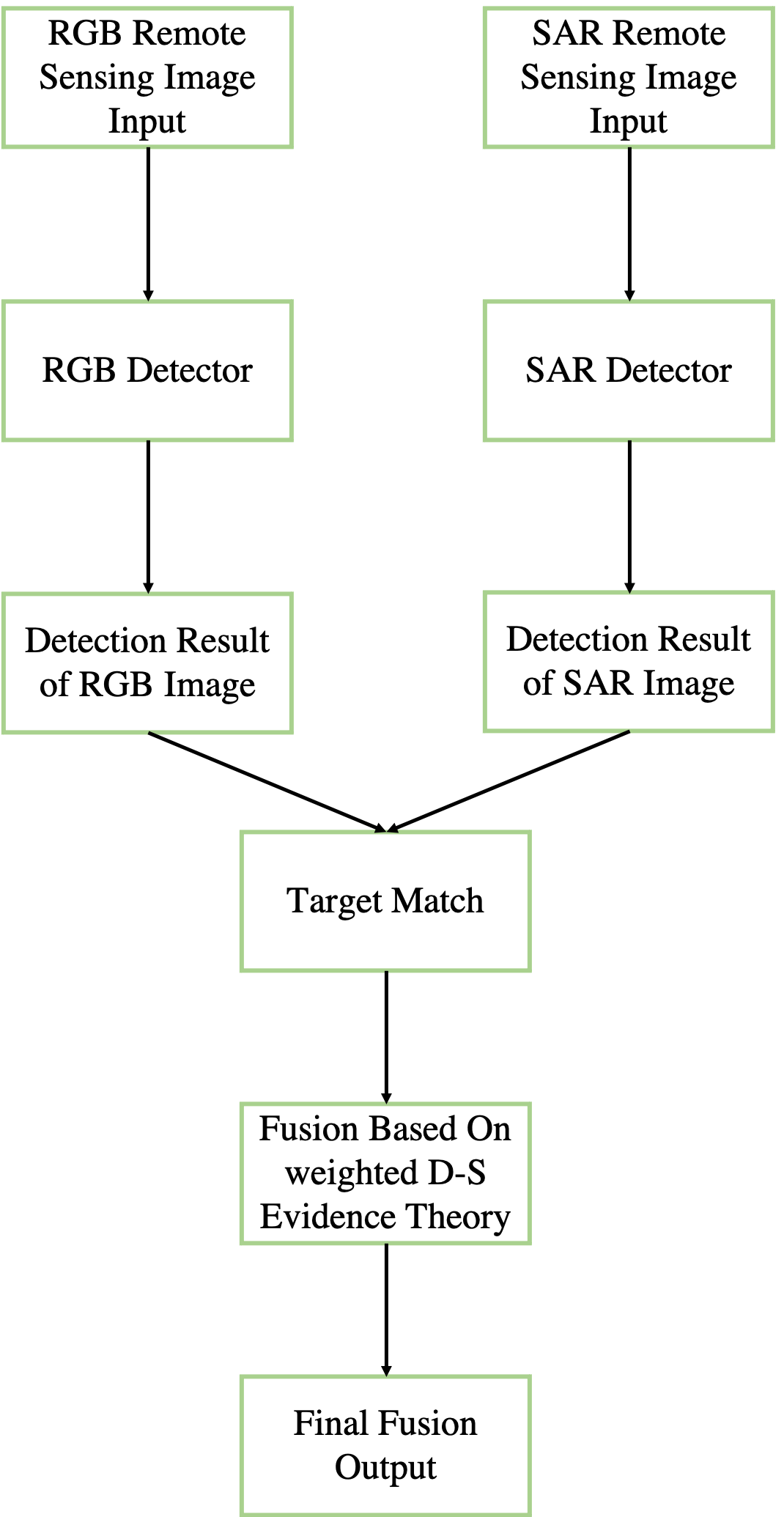} 
	\caption{Algorithm structure} 
	\label{algorithm}
\end{figure}

The framework of the fusion detection in multi-source remote sensing imagery is shown in Fig. \ref{algorithm}. Notably, it is the framework for fusion detection inference that is not included the training process. In terms of the inference there are following four steps:

\begin{itemize}
    \item A pair of roughly matched RGB image, i.e., optical image, and SAR image inputs.
    \item Detect RGB and SAR images by respective detectors.
    \item Perform fine target matching on the detection results.
    \item Fusion outputs based on weighted D-S evidence theory.
\end{itemize}

The target matching and fusion steps are described in detail in subsections C and D.

\begin{figure}[!htbp]
	\centering
	\includegraphics[width=0.7\textwidth]{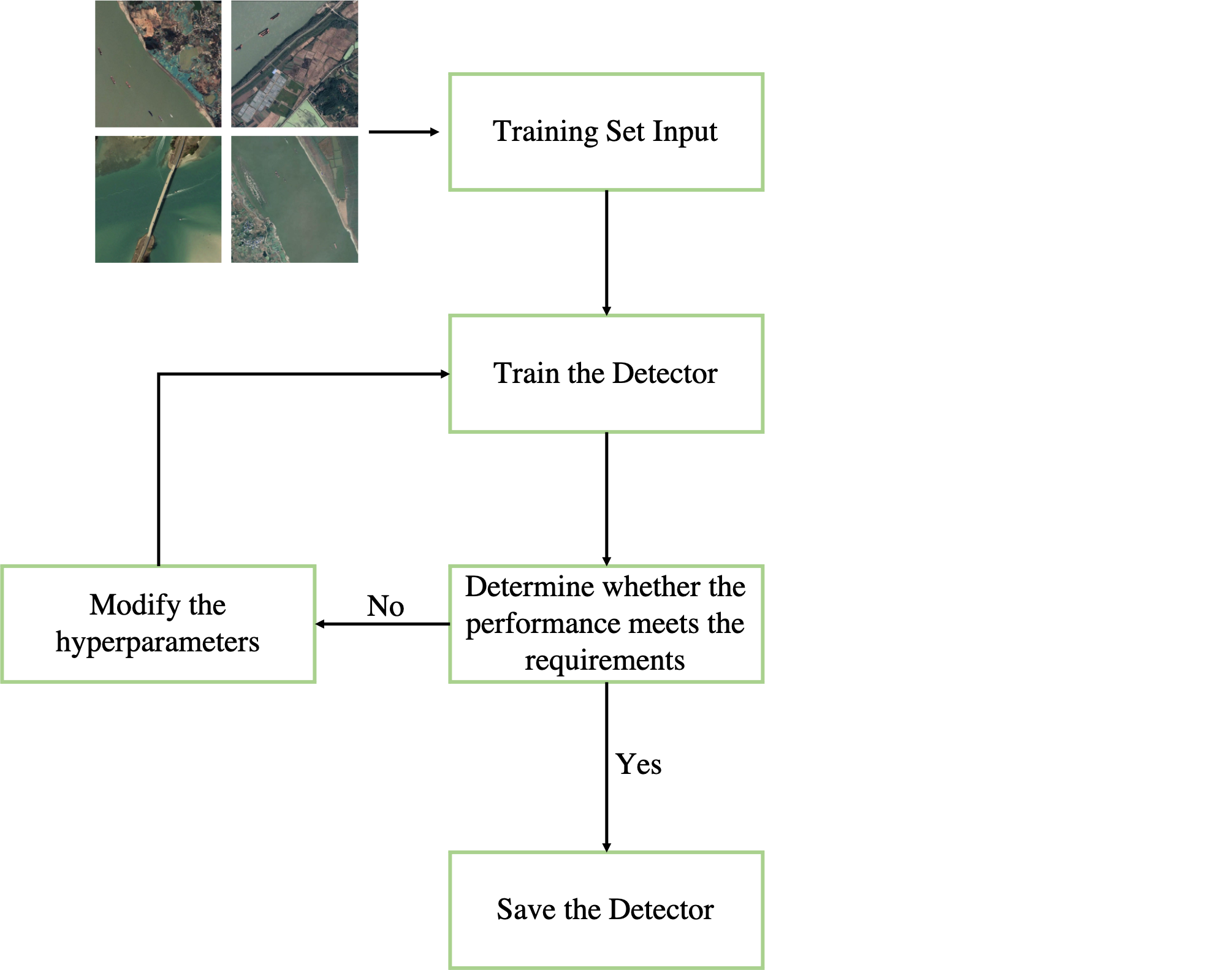} 
	\caption{Training process} 
	\label{train}
\end{figure}

Figure \ref{train} shows the training process of the detector, taking the RGB detector as an example. The detector is trained by the training sets.
There is a loop condition judgment, which means that the hyperparameters of the detector need to be adjusted if the performance is not met.
And the training process needs to run again until the performance is met. The SAR detector training process is the same, except that the RGB images training sets are replaced with the SAR image training sets.

\subsection{Strong Detector}
In the proposed fusion detection framework, the performance of the detector is critical, it directly affects the final detection accuracy. The YOLOv5 \cite{glennjocher20204154370} algorithm in the YOLO family is adopted as a strong detector in the current paper.

YOLO has attracted much attention since it was proposed and has become one of the typical one-stage object detection algorithms. Compared with the two-stage detectors, the most obvious advantage of YOLO is that it is simple and fast. Because of its extremely fast speed, it is widely applied in the industry. Compared with the previous version, the latest YOLOv5 has the following advantages:

\begin{itemize}
    \item There are four different scale models of super large, large, medium, and small to meet different needs.
    \item The accuracy is higher and the speed is faster. 
    \item With pyramid attention network  (PAN) \cite{li2018pyramid} and mosaic data augmentation, the detection performance of small objects is significantly improved.
\end{itemize}

In order to verify the strong performance of YOLOv5, some experiments are conducted on the HRSC2016 data set\cite{liu2016ship} which is an optical remote sensing image data set for ships and  AIR-SARship-2.0 data set\cite{xian2019air} which is a SAR remote sensing image data set for ships. The experimental results are shown in the Tab. \ref{tab:resultone}. The image size of the training set and the test set are both 640×640.

\begin{table}[!hbtp]
\caption{ Experimental results}

{\centering
\begin{tabular}{lccccccccc}
\hline
 & \multicolumn{2}{c}{HRSC2016} &  & \multicolumn{2}{c}{AIR-SARship-2.0}  \\ \cline{2-3}  \cline{5-6}
Model & ${\text{AP}_{0.50}}^{1}$  & ${\text{AP}_{0.50:0.05:0.95}}^2$  & &  ${\text{AP}_{0.50}}$  & $\text{AP}_{0.50:0.05:0.95}$ & Layers  & FPS$^3$,Hz& Params$^4$,M & FLOPs$^5$,G \\\hline
YOLOv5s$^6$ & 0.9610 & 0.8559 &  & 0.7875 & 0.4268 & 224 & 65.91 & 7.1 & 16.3\\
YOLOv5m$^7$ & 0.9666 & 0.8841 &  & 0.8165 & 0.4640 & 308 & 50.35 & 21.1 & 50.3\\
YOLOv5l$^8$ & 0.9601 & 0.8918 &  & 0.8058 & 0.4644 & 392 & 40.34 & 46.6 & 114.1\\
\hline
\end{tabular}
\label{tab:resultone}
}

${\text{AP}_{0.50}}^{1}$ is the PASCAL VOC metric, which means Average Precision(AP) at IoU=0.50.

${\text{AP}_{0.50:0.05:0.95}}^2$ is the primary challenge metric, which means AP at IoU=0.50:0.05:0.95. It is much more strict than  ${\text{AP}_{0.50}}$.

FPS$^3$ means Frames Per Second reflecting the detection speed.

Params$^4$ means Parameters, and M represents Mega.

FLOPs$^5$  means Floating Point Operations, and G means Giga.

YOLOv5s$^6$  means the small size model.

YOLOv5m$^7$  means the middle size model.

YOLOv5l$^8$  means the large size model.

\end{table}

It is noteworthy that we did not test YOLOv5x (super-large size) because we found that the YOLOv5l already showed overfitting. For single-category detectors, an overly complex structured network is redundant. In the optical and SAR image test sets, YOLOv5s can achieve extremely high accuracy, even when compared to the YOLOv5l, its parameters are reduced by 6.6 times, the FLOPs are reduced by 7 times. Only under stricter metric ${\text{AP}_{0.50:0.05:0.95}}$ can YOLOv5l outperform others. Therefore, considering the trade-off of speed and performance, YOLOv5s is adopted as the strong detector of ship detection in multi-source remote sensing images. 

Undeniably, a network with fewer parameters and smaller FLOPs can alleviate the burden on edge devices. In the future, it will be easier to transplant to the satellite on-board processor to achieve on-orbit object detection, thereby enhancing the intelligence level of the satellite.

%
\subsection{Target Matching based on DDIoU}

In most practical scenarios, the optical and SAR images come from different satellite platforms, and the shooting time of different remote sensing satellites is hard to be consistent. Even after the Spatio-temporal alignment of heterogeneous images, it is difficult to obtain a complete one-to-one correspondence of image pairs. When performing a decision-level fusion, it is necessary to consider the offsets of targets caused by factors such as sensor working time and Spatio-temporal matching errors. Therefore, A target matching for multi-source remote sensing images is needed to be considered.

The similarity is a representative metric for judging whether two targets are the same. What's more, the construction of the similarity function will be the key to the target matching. Most of the previous efforts mostly have adopted distance or weighted distance, such as Euclidean distance ($L_2$ norm), Manhattan distance ($L_1$ norm) or other weighted forms. Here is a kind of similarity based on the weighted Euclidean distance, formulaically

\begin{equation}
s\left(A, B\right)=\alpha_{1} \sqrt{\left(x_{1}-x_{2}\right)^{2}+\left(y_{1}-y_{2}\right)^{2}}+\alpha_{2} \sqrt{\left(w_{1}-w_{2}\right)^{2}+\left(h_{1}-h_{2}\right)^{2}}
\label{similarity}
\end{equation}

 where, $s(A,B)$ represents the similarity of the two targets, $(x_1, y_1, w_1, h_1)$ and $(x_2, y_2, w_2, h_2)$ are the x-coordinates, y-coordinates, width and length of the two targets, respectively. And $\alpha_1$ is the weight of the distance metric between the targets, and $\alpha_2$ is the weight of the length and width similarity between the two targets.

 However, the scale-variant feature is a major flaw in the similarity based on the distance measurement, that is, the size of two targets will greatly affect the final similarity value. Intersection over Union (IoU), also known as the Jaccard index, is the most commonly used metric for comparing the similarity between two arbitrary shapes.
 
 \begin{equation}
\operatorname{IoU}=\frac{A \cap B}{A \cup B}
\end{equation}

IoU encodes the shape properties of the target under comparison, e.g., the widths, heights, and locations, into the region property and then calculates a normalized measure that focuses on their areas, making IoU be scale-invariant.

Although IoU can effectively encode similarity, if two targets do not overlap, the IoU value will be zero. In the  non-overlapping case, the IoU value will be always 0, and can not reflect how far the two shapes are from each other. To solve this problem, a new IoU scheme is designed, which is called Distance-Decay Intersection over Union (DDIoU). The scheme introduces a distance metric, that is

\begin{equation}
\operatorname{DDIoU} = e^{-\alpha d} \cdot \operatorname{IoU}^{*}
\end{equation}

where $\operatorname{IoU}^{*}$ is a different version from original version , $d = \sqrt{\left(x_{1}-x_{2}\right)^{2}+\left(y_{1}-y_{2}\right)^{2}}$ and $\alpha$ is a weight of distance $d
$. $\operatorname{IoU}^{*}$ is obtained by calculating the IoU after forcibly aligning the center points of the two shapes.
Obviously, the $\operatorname{IoU}^{*}$ at this time only encode the shape information, and there is no distance information between the targets. The $e^{-\alpha d}$ function, as the distance decay term of DDIoU, introduces the distance information between two shapes into the similarity measurement. 
In this way, our proposed DDIoU  not only can encode the shape properties of the targets with scale-invariance but also can overcome the shortcoming that IoU is always equal to 0 when meeting non-overlapping cases, no matter how far the distance between the two targets is.

\begin{figure}[htbp]
	\centering
	
	\includegraphics[width=0.8\textwidth]{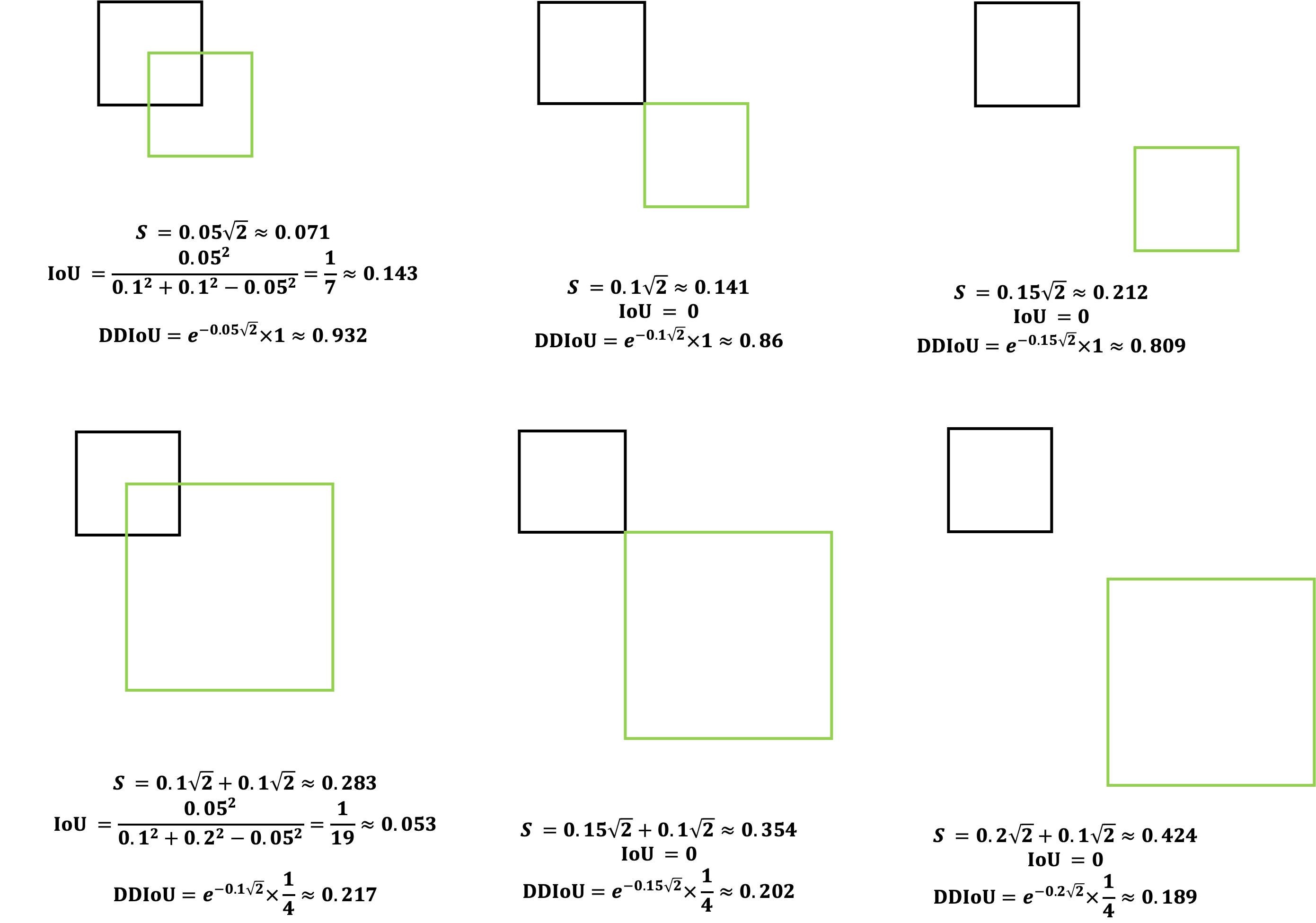} 
	\caption{Comparison of similarity indicators in six cases} 
	\label{camp}
	
\end{figure}

Figure \ref{camp} shows the comparison of similarities in six cases. For intuitive analysis, suppose $\alpha=1$, the two-color boxes are squares, and the box size is normalized relative to the image size. 
The side length of all black boxes is 0.1, and the side length of green boxes is only 0.1 in the first row and 0.2 in the second row. In addition, there are three cases in each row that are overlapped($d=0.05\sqrt{2}$),  just touched($d=0.1\sqrt{2}$)  and not touched($d=0.15\sqrt{2}$). For the convenience of calculation, the two weights of similarity $\alpha_1$ and $\alpha_2$ in formula \ref{similarity} are set to 1. It is worth noting that the smaller the value of $S(A,B)$, the more similar the two boxes are. When the value is equal to 0, it indicates that the two boxes are completely overlapped. Compared with $S(A,B)$, our proposed DDIoU pays more attention to the shape of the two boxes rather than the distance. That is, for two boxes with very similar shapes, DDIOU has a high tolerance for distance. Furthermore, taking the second case on the first line in Fig. \ref{camp} as an example, when the two boxes just touched and the side lengths were increased by 2 times, $S(A, B)$ also increased by 2 times($\frac{0.2\sqrt{2}}{0.1\sqrt{2}}=2$), while DDIoU only changed by 13.2\%($\frac{e^{-0.2\sqrt{2}}}{e^{-0.1\sqrt{2}}}\approx86.8\%$). It shows that compared with the former, DDIoU has the same property of normalization as IoU. However, IoU will fail when the two boxes do not overlap. DDIoU overcomes this shortcoming well due to the introduction of a distance-decay item. 

In general, DDIoU can not only retain the efficient coding capability of IoU for two arbitrary shapes but also overcome the defect of the inability to reflect the distance when the shapes do not overlap.

%
\subsection{Fusion based on weighted D-S Evidence Theory }


Frame of discernment and basic probability assignment(BPA) in the original D-S evidence theory are used in decision-level fusion, as follow,

\begin{itemize}
    \item[(1)] Frame of Discernment: assume $U$ is the universe, which represents the set of all possible states of the system under consideration. The frame of discrimination $\Theta$ is the power set of $U$, that is, the set of all subsets of $U$, 
    \begin{equation}
        \Theta = \mathscr{P}(U)
    \end{equation}
     where $\mathscr{P}(U)$ is the power set of $U$. The elements of frame $\Theta$ can be taken to represent propositions concerning the state of the system.
     
    \item[(2)] Basic Probability Assignment(BPA): assign a belief mass to each element of the power set. Formally, a function $m:\mathscr{P}(U) \rightarrow[0,1] $ is called a BPA. And it satisfies two properties. First, the mass function of the empty set is zero, i.e., $ m (\emptyset) = 0 $, and second, the sum of the masses of all members of $\Theta$ is 1, i.e., $\sum_{A\subset \Theta} m(A) = 1 $, where $A$ is a hypothesis.
    

    
    
\end{itemize}

    The fusion of evidence is obtained through Dempster's rule of combination, which is also the most critical operator in D-S evidence theory. For $\forall A \cap \Theta$, the combination of mass functions $m_1, m_2, \cdots, m_n$, i.e., the fusion output, are carried out in the following manner,

    \begin{equation}
        \begin{aligned}
            m_{{1,2,\cdots,n}}(\emptyset )&=0   \\
            m_{{1,2,\cdots,n}}(A ) = \left(m_{1} \oplus m_{2} \oplus \cdots \oplus m_{n}\right)(A) &= \frac{1}{K} \sum_{A_{1} \cap A_{2} \cap \cdots \cap A_{n}=A} m_{1}\left(A_{1}\right) \cdot m_{2}\left(A_{2}\right) \cdots m_{n}\left(A_{n}\right) 
        \end{aligned}
    \end{equation}

    where,
    \begin{equation}
        \begin{aligned}
            K &=\sum_{A_{1} \cap \cdots \cap A_{n} \neq \emptyset} m_{1}\left(A_{1}\right) \cdot m_{2}\left(A_{2}\right) \cdots m_{n}\left(A_{n}\right) \\
            &=1-\sum_{A_{1} \cap \cdots \cap A_{n}=\emptyset} m_{1}\left(A_{1}\right) \cdot m_{2}\left(A_{2}\right) \cdots m_{n}\left(A_{n}\right)
        \end{aligned}
    \end{equation}

If there is a high degree of conflict between evidence, counter-intuitive results will be generated by Dempster's rule\cite{zadeh1986simple}, which is Zadeh's paradox. In order to deal with this conflict, a weighted D-S evidence theory based on compatibility coefficients is employed in the current paper.

Suppose the frame of Discernment $\Theta=\left\{A_1,A_2,\cdots,A_M\right\}$, and $m_1,\ m_2,\ \cdots,m_n$ are mass functions, then the relative compatibility coefficient  of any two pieces of evidence $R_{ij}\left(A_k\right)$:

\begin{equation}
R_{ij}(A_{k})=\frac{2 m_{i}(A_{k}) m_{j}(A_{k})}{m_{i}(A_{k})^{2}+m_j(A_{k})^{2}}, k=1,2, \cdots, M
\end{equation}

When $m_i(A_k)=m_j(A_k)$ , the value of the compatibility coefficient equals 1, indicating that these two pieces of evidence agree on the hypothesis and there is no conflict. Meanwhile, the coefficient can also indicate the compatibility between evidence. 

The formula of the absolute compatibility coefficient is as follows,

\begin{equation}
D_{i}(A_{k})=\sum_{j=1}^{N} R_{i j}(A_{k})-1, j=1,2, \cdots, N
\end{equation}

The formula for the credibility of the hypothesis ,i.e. the weight of the hypothesis, is as follows,

\begin{equation}
w_{i}(A_{k})=\frac{D_{i}(A_{k})}{N-1}
\end{equation}

Finally, we can get the new weighted mass function $m_i^{\prime}(A_k)$ as follows,

\begin{equation}
m_{i}^{\prime}(A_{k})=w_{i}(A_{k}) m_{i}(A_{k})
\end{equation}

When there is conflict between the evidence, the corresponding weight of the individual evidence will be less than 1. 
Obviously, the deterministic information of the revised evidence will decrease, while the uncertainty of the hypothesis will increase, that is

\begin{equation}
m_{i}^{\prime}(\Theta)=1-\sum_{k=1}^{N} m_{i}^{\prime}(A_{k}), i=1,2, \cdots, n
\end{equation}

\begin{figure}[hbpt]
  \centering
  \subfigure[Optical Image]{\includegraphics[width=3in]{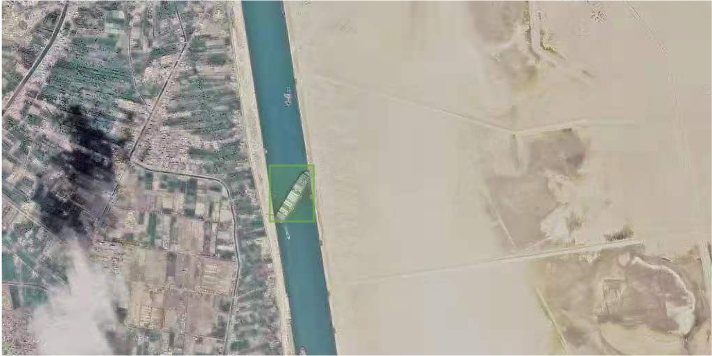}} \hspace{.2in}
  \subfigure[SAR Image]{\includegraphics[width=3in]{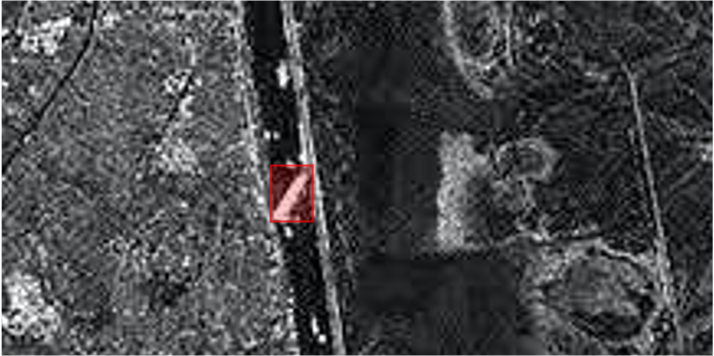}}
  \caption{ Container Ship EVER GIVEN Ran Aground in the Suez Canal }
  \label{fig:RGBandSAR}
\end{figure}

The paired optical and SAR images of container ship EVER GIVEN ran aground in the Suez Canal are taken as an example to show how our weighted D-S evidence theory can fuse the detection results of the two-source data in Fig.  \ref{fig:RGBandSAR}.

Assume that after target matching (the ship detection is marked with a green bounding box in the optical image, and the red bounding box is marked in the SAR image), the mass function of container ship EVER GIVEN to be detected in the optical image is 0.9, and the mass function of being detected in the SAR image is 0.8, as shown in Tab. \ref{tab:mass}. Then m$_1$ and m$_2$ represent detection in the optical image and SAR image, respectively, and hypothesis A$_1$ represents the ship exist, hypothesis A$_2$ means it does not exist. The final fusion results by weighted D-S evidence theory are shown in Tab. \ref{tab:DS-Fusion}, the probability of the ship EVER GIVEN being detected is 0.9725. It can be found that after fusion, the detection probability of the ship EVER GIVEN is increased, which helps to improve the performance of the detector.

\begin{table}[hbt!]
\caption{Mass function}
\centering
\begin{tabular}{ccc}
\hline
Hypothesis & m$_{1}$  & m$_{2}$ \\
\hline
A$_1$   & 0.9 & 0.8 \\ 
 A$_2$   & 0.1 & 0.2 \\
\hline
\end{tabular}
\label{tab:mass}
\end{table}

\begin{table}[hbt!]
\caption{The fusion result of weightd D-S evidence theory}
\centering
\begin{tabular}{cccc}
\hline
Hypothesis & m$_{1}^{\prime}$  & m$_{2}^{\prime}$ & Fusion (m$_{1,2}^{\prime}$) \\
\hline
 A$_1$   & 0.8938 & 0.7945 & 0.9725\\ 
 A$_2$   & 0.0800 & 0.1600 & 0.0260\\
$\Theta$ & 0.0262 & 0.0455 & 0.0015\\
\hline
\end{tabular}
\label{tab:DS-Fusion}
\end{table}

\section{Performance Experiment}

\subsection{Dataset and Experiment Settings}

\begin{figure}[hbpt]
  \centering
  \subfigure[Optical image in the training set]{\includegraphics[width=3in]{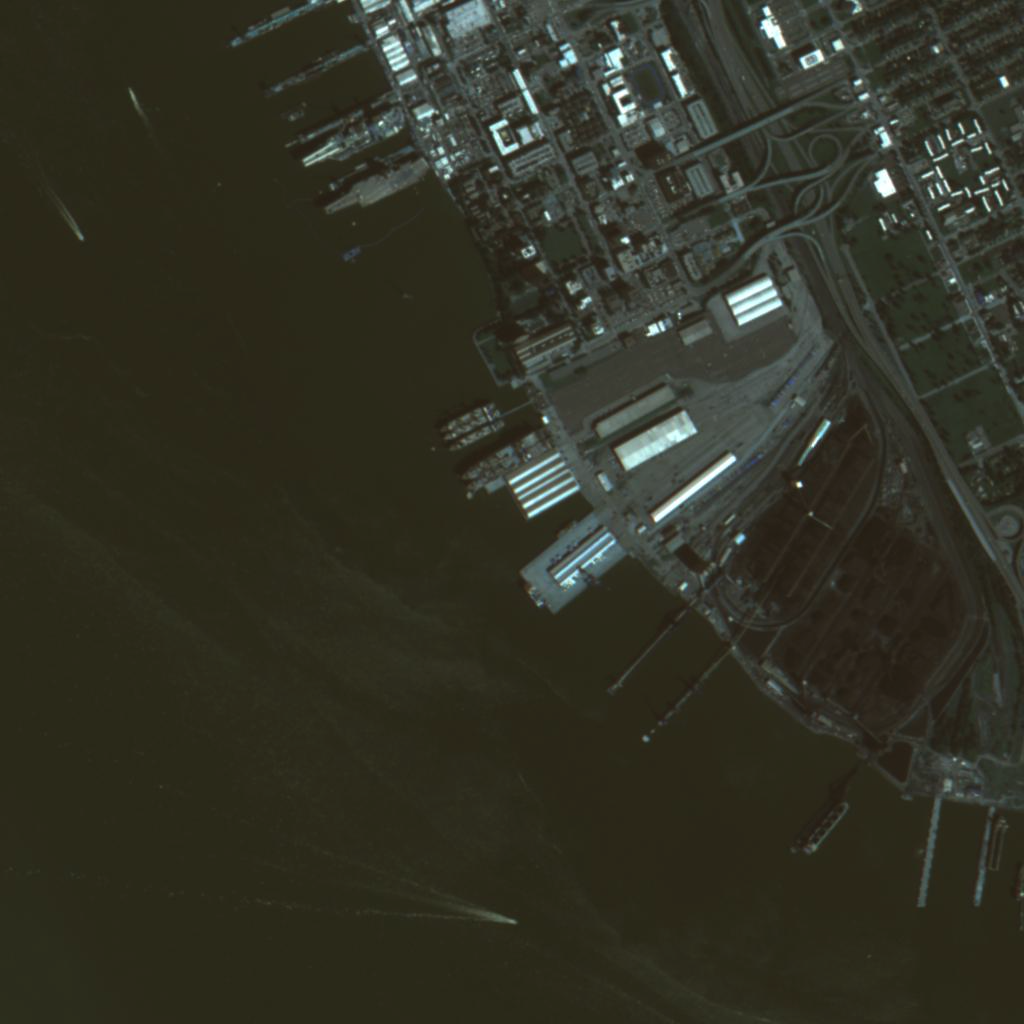}} \hspace{.2in}
  \subfigure[SAR image in the training set]{\includegraphics[width=3in]{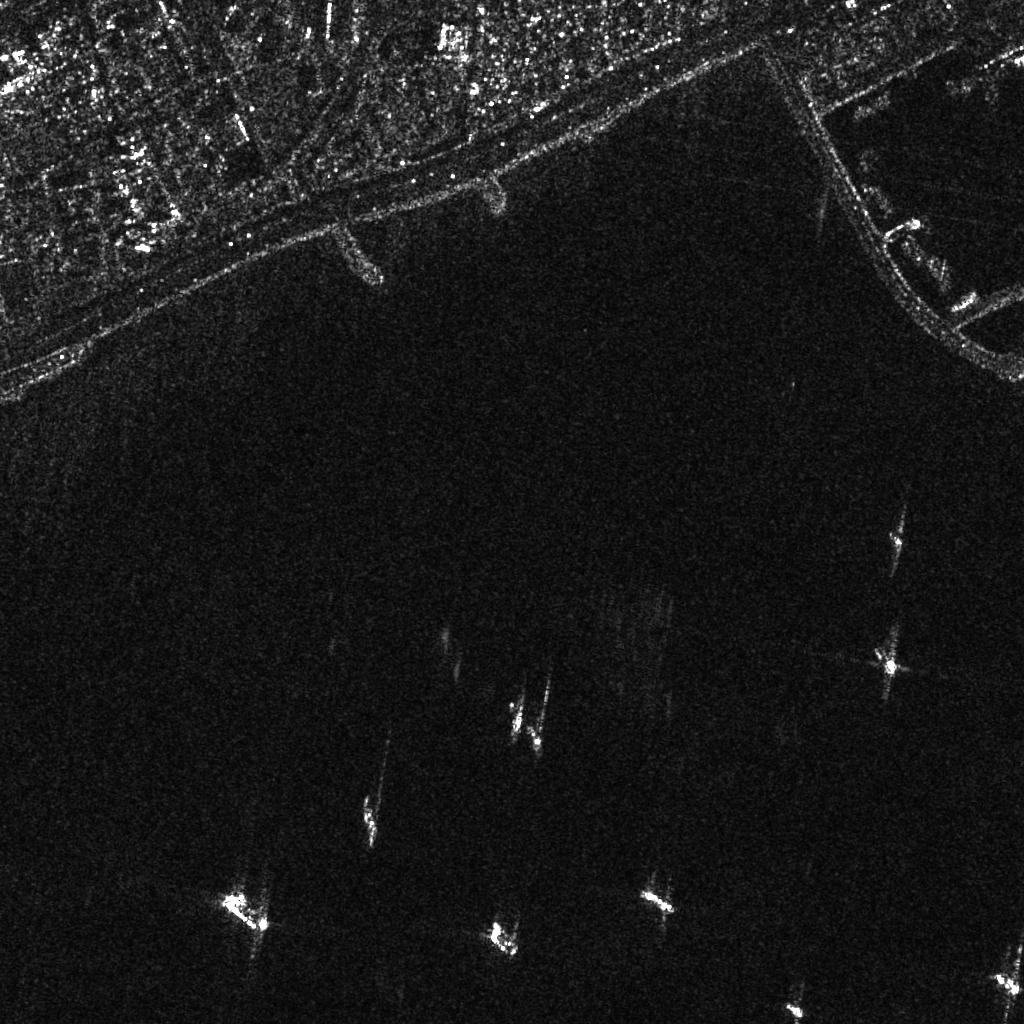}}
  \subfigure[Optical image in the paired test set]{\includegraphics[width=3in]{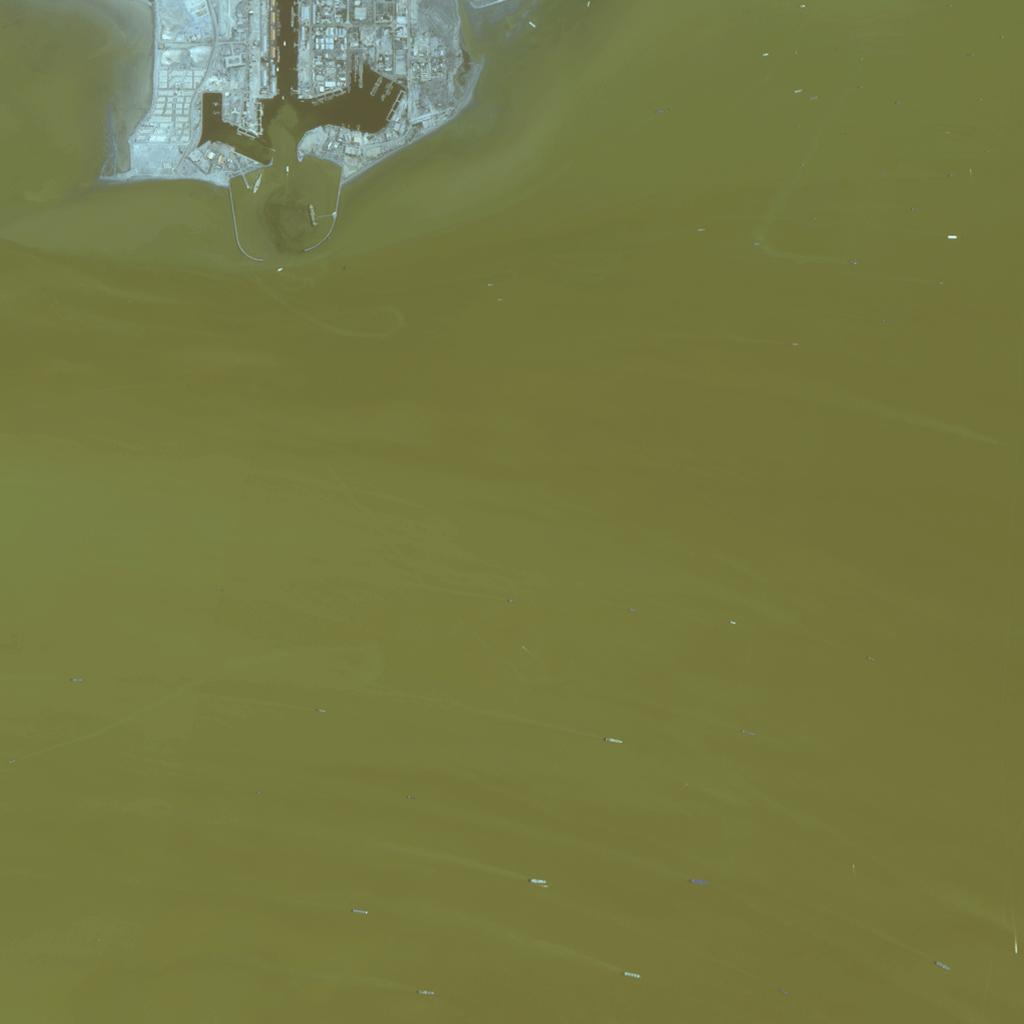}} \hspace{.2in}
  \subfigure[SAR image in the paired test set]{\includegraphics[width=3in]{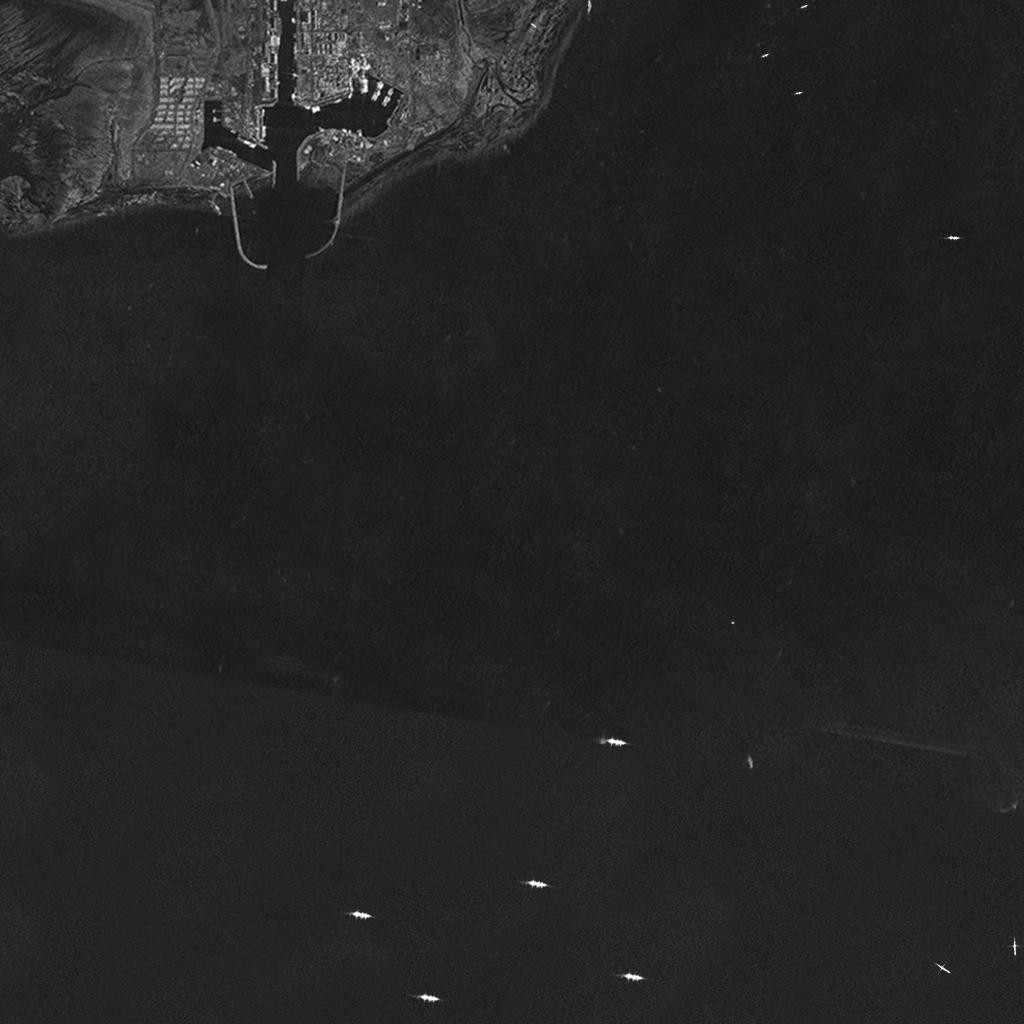}}
  \caption{Examples of training set and test set }
  \label{fig:examples}
\end{figure}

A data set of the optical and SAR remote sensing images is constructed, including 282 optical images and 281 SAR images as training sets, and 76 pairs of optical and SAR paired images as test sets. The test set contains a total of 1152 ship targets. 
It is worth to be noted that because it is difficult to obtain optical and SAR images at the same time, images that are taken at the same place but nonidentical time were selected for the test set. 
To pair the images from different sources, a ship which can be observed in the optical images but not in SAR images was added artificially in the SAR images.
The added ship has the same position as seen in the optical image.
Examples for the training set and test set are presented in Fig.\ \ref{fig:examples}.

The software and hardware platforms for training and inference of fusion detection are configured as follows, Intel(R) Core(TM) i9-7900X CPU@3.3GHz (CPU), 2 × NVIDIA$^\text{®}$ TITAN RTX™  24G(GPU), 4 × DDR4 8G (Memory), Samsung 960 Pro 512G (Hard disk), Ubuntu18.04 LTS (System), and Pytorch(Open source framework) with CUDA11.0, cudnn8.1.

 All detectors are trained by using the Adam algorithm, where the initial learning rate is 1e-3. The total epoch is 300, and the batch size is 32 for all training. And the input size for both training and test images is 640×640.

\subsection{Evaluation Indicators}
In order to measure the performance of fusion detection in multi-source remote sensing imagery, Precision, Recall and $\text{AP}_{0.5}$ three indicators are adopted in this paper.

Precision, Recall, and AP are shown in the following three formulas respectively, 

\begin{equation}
    \text { Precision }=\frac{TP}{TP+FP} 
\end{equation}
\begin{equation}
\text { Recall }=\frac{TP}{TP+FN}
\end{equation}
\begin{equation}
\text{AP}=\int_{0}^{1}  \text { Precision } d \text { Recall }
\label{equ:ap}
\end{equation}

Where $TP$ is true positive, which means that the prediction bounding box by detection and the ground truth meet the IoU threshold; otherwise, it will be considered as false positive ($FP$). Flase negative ($FN$) means there is a true target, but detector misses it.  Equation \ref{equ:ap} shows that $\text{AP}$ is the integral of the Precision-Recall Curve (PRC). And $\text{AP}_{0.5}$ means AP at IoU=0.50.

\subsection{Performance Result}

\begin{figure}[hbpt]
  \centering
    \subfigure[Results of the optical detector on the optical image]{\includegraphics[width=2.5in]{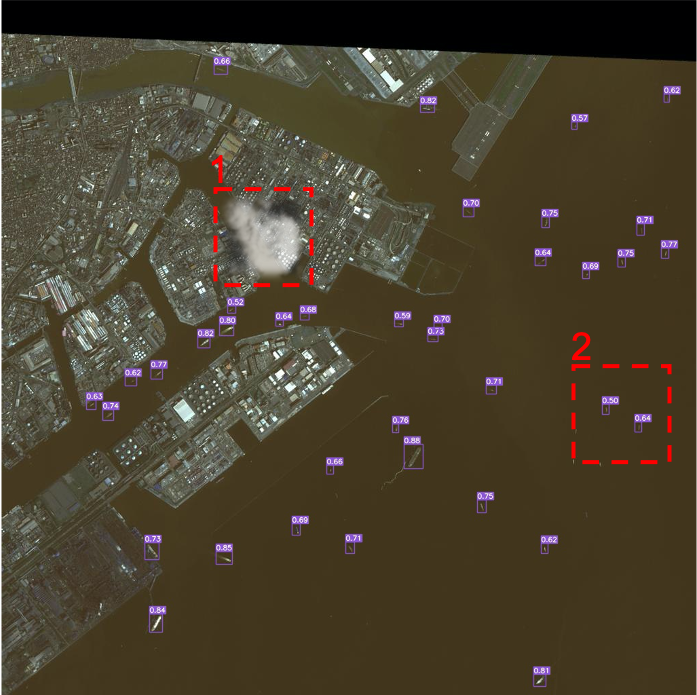}} \hspace{.1in}
    \subfigure[Results of the SAR detector on the SAR image]{\includegraphics[width=2.5in]{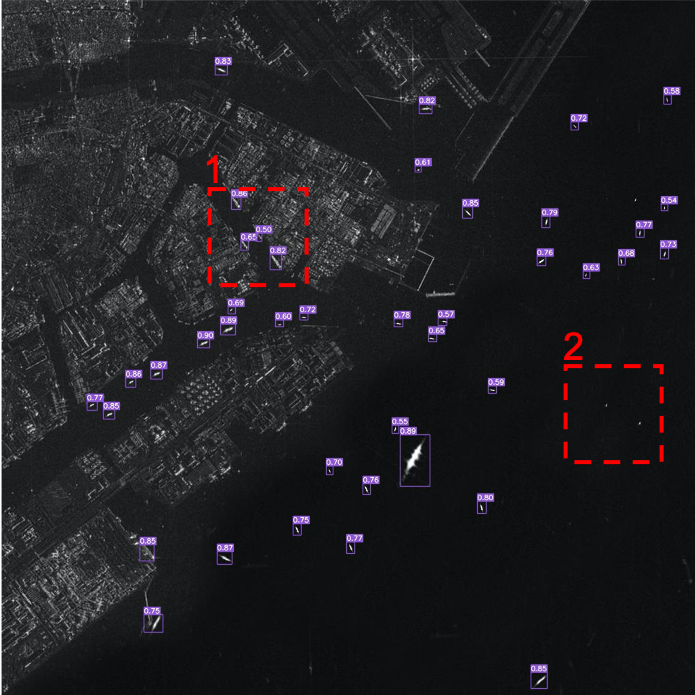}}
    \subfigure[Results of the DDIoU-based fusion detector on the optical  image]{\includegraphics[width=2.5in]{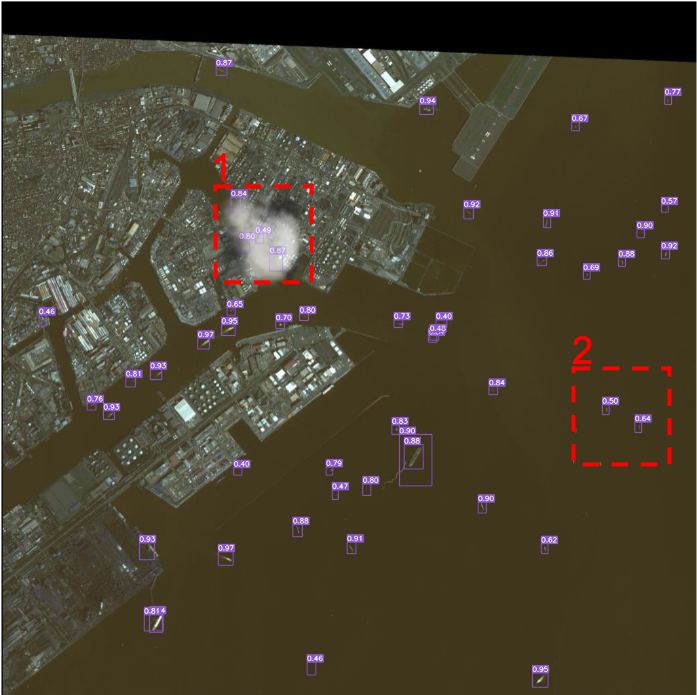}} \hspace{.1in}
    \subfigure[Results of the DDIoU-based fusion detector on the SAR  image]{\includegraphics[width=2.5in]{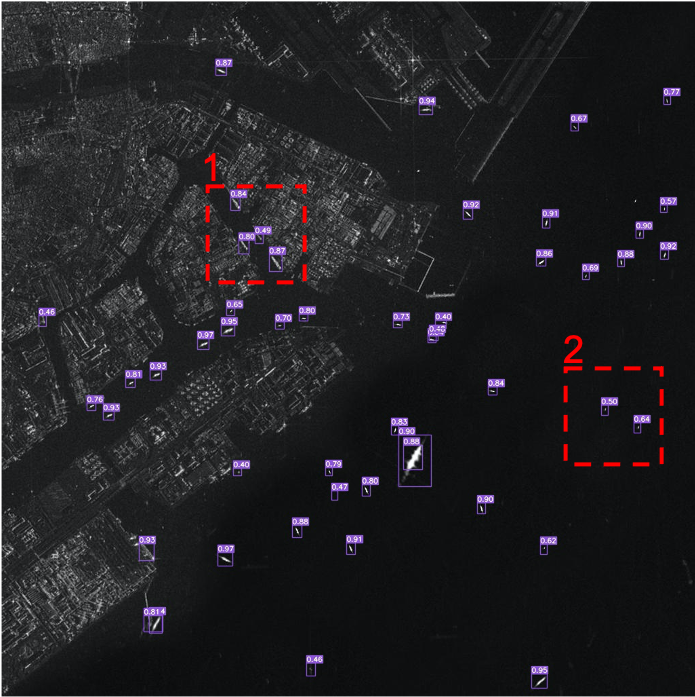}}
    \subfigure[3 times magnified area 1]{\includegraphics[width=2.5in]{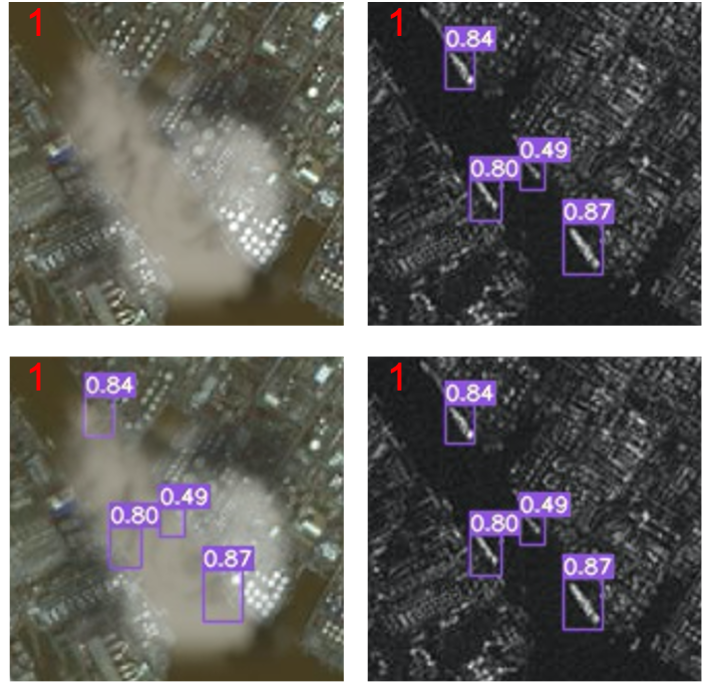}} \hspace{.1in}
    \subfigure[3 times magnified area 2]{\includegraphics[width=2.48in]{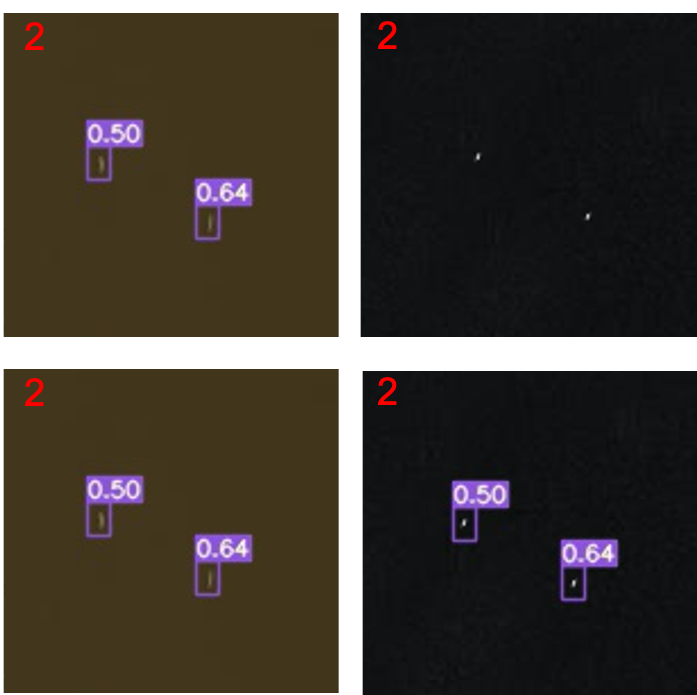}}
  \caption{Visualization of detection results by optical detector, SAR detector  and fusion detector}
  \label{fig:fusionDet}
\end{figure}

To demonstrate that our proposed fusion detection framework is powerful, Experiments on the RGB and SAR paired test sets are conducted.
In order to better illustrate, an example detection result of the test set is visualized in Fig. \ref{fig:fusionDet}.
In the following example, the above Fig. (\ref{fig:fusionDet}, a)-(\ref{fig:fusionDet}, d) are the detection results of the optical detector on the optical images, the detection results of the SAR detector on the SAR images, and the detection results of the DDIoU-based fusion detector on the optical and SAR images, respectively. 
And for each test result, two areas that are bounded by red boxes and marked with numbers should be noted.
For a more intuitive comparison, area 1 and area 2 have been enlarged 3 times in Fig. (\ref{fig:fusionDet}, e) and Fig. (\ref{fig:fusionDet}, f). 
For area 1, due to the occlusion of clouds, the ship target cannot be detected by the optical detector in Fig. (\ref{fig:fusionDet}, a) and the upper left corner of Fig. (\ref{fig:fusionDet}, e).
Conversely, the nature of SAR imaging allows the SAR detector to detect ship targets under clouds very well.
In Fig. (\ref{fig:fusionDet}, c), (\ref{fig:fusionDet}, d) and (\ref{fig:fusionDet}, e), it can be found that the proposed detection method overcomes the shortcoming that missed detection caused by cloud occlusion by means of fusion.
Besides, for area 2, the SAR image has less information than the optical image, while the lack of information can lead to failure in detecting the vessel in Fig. (\ref{fig:fusionDet}, b) and the upper right corner of Fig. (\ref{fig:fusionDet}, f). 
On the contrary, the richer details for the detection can be provided in the optical image to achieve more accurate detection in Fig. (\ref{fig:fusionDet}, a). Furthermore, the problem of weak detection in SAR images is alleviated by fusing the optical information in Fig. (\ref{fig:fusionDet}, a). In short, our proposed fusion detection method can fully combine and enhance the cross-modal information between optical and SAR images to detect ship targets more accurately.

\begin{figure}[hbpt]
  \centering
\includegraphics[width=0.6\textwidth]{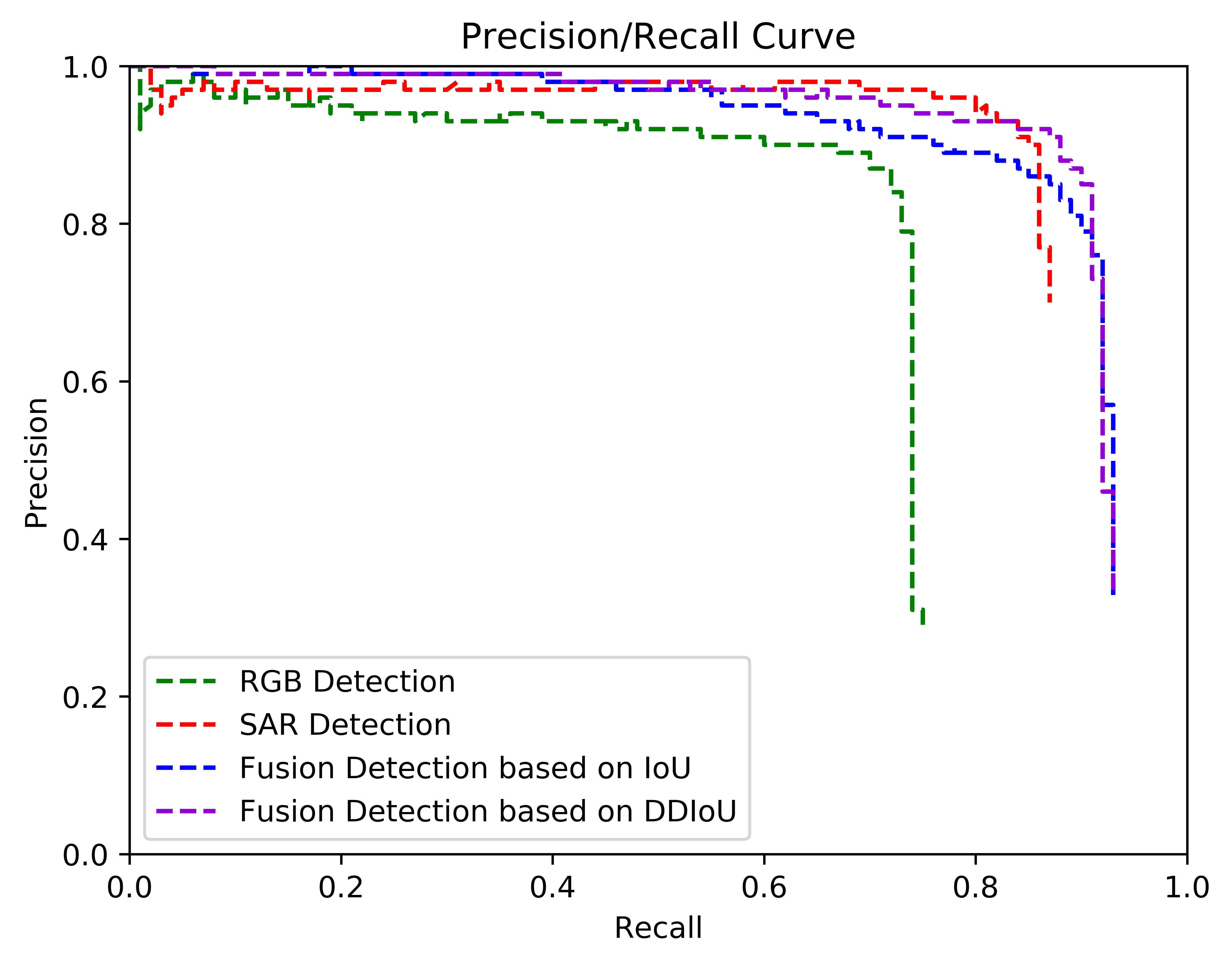}
  \caption{ Precision-Recall Curves }
  \label{fig:Pre-rec_curve}
\end{figure}

 Precision-Recall Curves(PRCs) are drawn to further prove the robustness and efficiency of our proposed algorithm. The curve at the top right of the PRCs indicates a better performance. Figure \ref{fig:Pre-rec_curve} shows the PRCs of two fusion detection methods as well as RGB and SAR detection. $\text{AP}_{0.5}$ , FPS and  Fusion Time, that is, required for running fusion algorithm, are shown in Tab. \ref{tab:result}, where the best value of each item is indicated in bold.  Compared with RGB detection and SAR detection, our proposed fusion detection based on  DDIoU and weighted D-S evidence theory has increased by 20.13\% and 4.95\% on  $\text{AP}_{0.5}$  respectively. By jointly analyzing the $\text{AP}_{0.5}$ and the PRCs, our proposed method shows superior performance for ship targets in multi-source remote sensing imagery. Although the two fusion detection methods are relatively slow, their detection speed can also reach near-real-time (30Hz).  In addition, the execution time of the fusion algorithm is only 4-5 milliseconds, which is very small compared to the whole detection time. This also indicates that our decision-level fusion algorithm is lightweight and does not impose too much computational burden on the device, i.e., the algorithm is device-friendly. It is crucial to on-orbit multi-source detection on satellites in the future.


\begin{table}[!hbpt]
\caption{Performance Results of four methods}
\centering
\begin{tabular}{cccc}
\hline
Methods & $\text{AP}_{0.5}$  & FPS,Hz & Fusion Time, ms\\
\hline

 RGB Detection      				     & 0.6931    &  \textbf{65.91}   &  /   \\ 
 SAR Detection       				     & 0.8449    &  \textbf{65.91}   &  /    \\
Fusion Detection\textsuperscript{*}   & 0.8829    &  28.78   &  \textbf{4.40}\\
Fusion Detection\textsuperscript{†}  & \textbf{0.8944} &  28.29   &  5.00\\
\hline
\end{tabular}
\label{tab:result}

Fusion Detection\textsuperscript{*} is the fusion detection  based on IoU and weighted D-S evidence theory.\\
Fusion Detection\textsuperscript{†} is the fusion detection  based on DDIoU and  weighted D-S evidence theory.
\end{table}

\section{Conclusion}

In order to fully leverage the cross-modal information between multi-source remote sensing images, a fast detection method based on DDIoU and weighted D-S evidence theory for decision-level fusion is designed. Compared to traditional single-source detection, e.g. optical or SAR image, target perception by current method is more comprehensive, which facilitate a better visual interpretation. The weighted D-S evidence theory is a new decision-level fusion method, which is based on the compatibility coefficients between the evidences. And the drawback of feature-level fusion that requires a large amount of paired training data is overcome by the weighted D-S evidence theory. More importantly, the fusion algorithm is lightweight and takes only 4 to 5 milliseconds to run. The DDIoU scheme introduces a distance-decay term into IoU, which not only retains the effective coding capability of IoU for two arbitrary shapes, but also overcomes the defect that the distance cannot be reflected in the non-overlapping case. Therefore, the robustness and flexibility are improved for target matching. In the experiments carried out in present paper, the $\text{AP}_{0.5}$ of the proposed method is 89.44\%, which is 20.13\% higher than optical detection and 4.9\% higher than SAR detection.

\section*{Funding Sources}


This work was supported by the National Natural Science Foundation of China under Grant No.U20B2056  and the Key Laboratory of Intelligent Infrared Perception, Chinese Academy of Sciences.


\bibliography{sample}

\end{document}